%% file: arxiv_errata.tex
\documentclass[10pt,twocolumn,letterpaper]{article}

\usepackage[pagenumbers]{cvpr} % To force page numbers, e.g. for an arXiv version
\usepackage{soul}
\usepackage{color,xcolor}
\usepackage{graphicx}
\usepackage{amsmath}
\usepackage{amssymb}
\usepackage{booktabs}
\usepackage{multirow}
\input{definitions}
\usepackage[accsupp]{axessibility}

\usepackage[pagebackref,breaklinks,colorlinks]{hyperref}

\usepackage[capitalize]{cleveref}
\crefname{section}{Sec.}{Secs.}
\Crefname{section}{Section}{Sections}
\Crefname{table}{Table}{Tables}
\crefname{table}{Tab.}{Tabs.}

\def\cvprPaperID{2627} % *** Enter the CVPR Paper ID here
\def\confName{CVPR}
\def\confYear{2023}

\begin{document}

\title{TRACE: 5D Temporal Regression of Avatars \\with Dynamic Cameras in 3D Environments\vspace{-4mm}}
\makeatletter
\g@addto@macro\@maketitle{
\centerline{
  \includegraphics[width=0.92\textwidth]{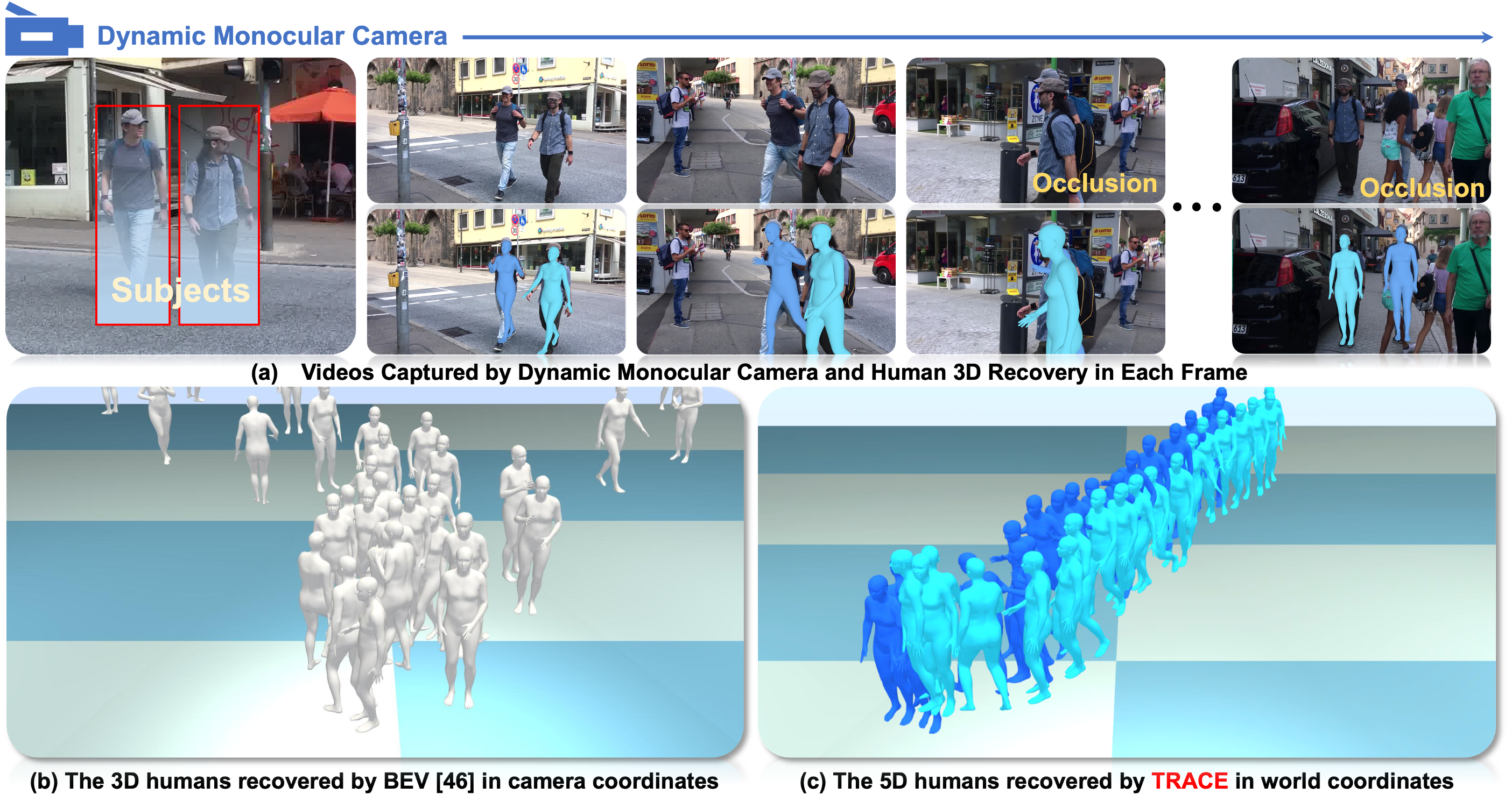}}
  \vspace{-0.1in}
\captionof{figure}{\textbf{5D temporal reconstruction of multiple 3D people in global coordinates with a dynamic camera.} We introduce TRACE, a monocular one-stage method with a holistic 5D representation that enables the network to explicitly reason about human motion across time. Unlike previous methods~\cite{bev,romp} \textbf{(b)} that regress all 3D people from each frame in camera coordinates, TRACE \textbf{(c)} tracks the subjects \textbf{(a)} presented in the first frame through time and recovers their global trajectories in global coordinates in one shot.  }\vspace{5mm} %\leftline{}
\label{fig:teaser}
}

\author{Yu Sun$^1$\thanks{This work was done when Yu Sun was an intern at JD.com. }\quad
Qian Bao$^{2\dag}$ \quad
Wu Liu$^{2}$\thanks{Corresponding author.} \quad
Tao Mei$^3$\quad
Michael J. Black$^4$ \quad \\
$^1$Harbin Institute of Technology \quad $^2$Explore Academy of JD.com \\
$^3$HiDream.ai Inc. \quad $^4$Max Planck Institute for Intelligent Systems \\
{\tt\small \texttt{yusunhit@gmail.com, baoqian@jd.com, liuwu1@jd.com}}\\
{\tt\small\texttt{tmei@hidream.ai, black@tuebingen.mpg.de}}\vspace{-7mm}
}
%, T\"ubingen, Germany

\maketitle
\input{Sections/abstract}
\input{Sections/introduction}
\input{Sections/related_work}
\input{Sections/method}
\input{Sections/experiments}
\input{Sections/conclusions}
\input{Sections/acknowledgements}
\input{Sections/errata}

%%%%%%%%% REFERENCES
{\small
\bibliographystyle{ieee_fullname}
\bibliography{arxiv_errata}
}

\end{document}

%%%%%%%%% PAPER ID  - PLEASE UPDATE
\def\cvprPaperID{2627} % *** Enter the CVPR Paper ID here
\def\confName{CVPR}
\def\confYear{2023}

\title{TRACE: 5D Temporal Regression of Avatars \\with Dynamic Cameras in 3D Environments\\ \textbf{**Supplementary Material**}}
\maketitle
\section{Introduction}

In this supplemental document, we provide more implementation details and discuss the limitation of TRACE. 
Please refer to the \textbf{supplemental video} for more qualitative results and video sequences.

\section{Implementation Details}

In this section, we introduce the details of our loss functions, temporal feature propagation module, inference and training details, and dataset details.

\subsection{Loss Functions}
Except for the temporal motion losses we introduced in the main paper, we also follow previous work~\cite{romp,hmr,bev} to supervise the estimated maps and SMPL parameters with standard image losses.
$\boldsymbol{\mathcal{L}}_{cm}$ is the focal loss~\cite{romp} of the estimated 3D Center map.
We also supervise the focal loss of the 2D Body Center heatmap, which is used for BEV-based 3D composition of the 3D Center map.
%In the same pattern, we further use a 3D focal loss $\boldsymbol{\mathcal{L}_{cm3D}}$ to supervise the 3D Center map via converting $\boldsymbol{\mathcal{L}_{cm}}$'s 2D operation to 3D.
The SMPL parameter loss $\boldsymbol{\mathcal{L}}_{pm}$ consists of four parts, $\boldsymbol{\mathcal{L}_{\theta}},\boldsymbol{\mathcal{L}_{\beta}}$, $\boldsymbol{\mathcal{L}}_{prior}$, and $\boldsymbol{\mathcal{L}_{J}}$.
$\boldsymbol{\mathcal{L}_{\theta}}$ and $\boldsymbol{\mathcal{L}_{\beta}}$ are $L_2$ losses on SMPL pose $\boldsymbol{\theta}$ and shape $\boldsymbol{\beta}$ parameters respectively.
In the first 20 epochs of training, we also employ the Mixture of Gaussian pose prior~\cite{keep,smpl}, $\boldsymbol{\mathcal{L}}_{prior}$, to penalize the unreasonable poses $\boldsymbol{\theta}$.

The 3D body keypoint loss $\boldsymbol{\mathcal{L}_{J}}$ consists of 3 parts, $\boldsymbol{\mathcal{L}}_{mpj}$, $\boldsymbol{\mathcal{L}}_{pmpj}$, and $\boldsymbol{\mathcal{L}}_{pj2d}$.
$\boldsymbol{\mathcal{L}}_{mpj}$ is the $L_2$ loss of 3D body keypoints $\boldsymbol{J}$.
Following~\cite{romp,sun2019dsd-satn,bev}, we employ $\boldsymbol{\mathcal{L}}_{pmpj}$ to alleviate the domain gap between training datasets, which is the $L_2$ loss of the predicted 3D body keypoints $\boldsymbol{J}$ after Procrustes alignment with the ground truth.
We also employ $\boldsymbol{\mathcal{L}}_{pj2d}$ to learn from 2D pose datasets, which is the $L_2$ loss of the 2D projection $\boldsymbol{J}_{2D}$ of 3D joints $\boldsymbol{J}$ using predicted 3D position $\boldsymbol{\widehat{t}_{i}}$ in camera coordinates.

\subsection{Temporal Feature Propagation Module}
To extract long-term and short-term motion features, we construct a temporal feature propagation module by combining a ConvGRU~\cite{teed2020raft} module, a Deformable convolution~\cite{zhu2019deformable} module, and a residual connection.
Inspired by~\cite{teed2020raft}, we employ a ConvGRU module to hold the long-term memory within a hidden state $\boldsymbol{H}_{i-1}$ and progressively update the memory with image feature map $\boldsymbol{F}_{i}$ to generate the new hidden state $\boldsymbol{H}_{i}$. 
Inspired by~\cite{bertasius2019learning}, to extract short-term motion features, we use $\boldsymbol{F}_{i}-\boldsymbol{F}_{i-1}$ to generate the feature sampling offset for Deformable convolution to process $\boldsymbol{F}_{i}$.
With these modules, we progressively fuse the image feature maps ($\boldsymbol{F}_{i-1}$, $\boldsymbol{F}_{i}$) to generate a temporal image feature map $\boldsymbol{F}^{\prime}_i$. 

During inference, we sequentially process every clip of a video sequence.
With a GRU-based temporal propagation module, the size of the video clips can be flexibly set.

\subsection{Inference and Training Details}

\textbf{Inference details.}
The input images are resized to $512\times512$.
The size of output maps is $D=64,H=128,W=128,C=128$.
The hyper-parameters of the memory unit are set to $\lambda_c=0.05,\lambda_s=0.13,\lambda_d=0.05,\lambda_m=1,W=(1.2,2.5,25),\lambda_f=100$.
Following~\cite{kolotouros2019spin,romp,bev}, we adopt the 6D representation~\cite{Zhou_2019_CVPR} for SMPL pose parameters $\boldsymbol{\theta}$. 
To improve the temporal smoothness, we also employ the One-Euro filter~\cite{casiez20121} to process the estimated SMPL pose parameters and 3D trajectories $\boldsymbol{\widehat{t}},\boldsymbol{T}$.

\textbf{Training details.}
The confidence threshold of the Body Center heatmap is 0.12.
The sampling ratios of different datasets are 18\% Human3.6M~\cite{h36m}, 20\% MPI-INF-3DHP~\cite{mehta2018single}, 14\% 3DPW~\cite{3dpw}, 12\% PennAction~\cite{pennaction}, 16\% PoseTrack~\cite{doering2022posetrack21}, 20\% DynaCam.
$w_{(.)}$ denotes the corresponding weight of all loss items.
We set these loss weights to $w_{m}=300,w_{W}=100,w_{mpj}=300,w_{cm}=100,w_{pmpj}=260,w_{pj2d}=300,w_{prior}=1.6,w_{\theta}=80,w_{\beta}=60$.

\subsection{Dataset Details}
In this section, we introduce the details of used datasets.
First, we introduce the 3D human pose datasets, 3DPW~\cite{3dpw}, Dyna3DPW~\cite{3dpw}, Human3.6M~\cite{h36m}, and MPI-INF-3DHP~\cite{mehta2018single}.

\begin{table}
	\begin{center}
			\begin{tabular}{l|l}
				\hline
				\textbf{Sequence Name}  & \textbf{Frame Ranges} \\
				\hline
				downtown\_warmWelcome\_00 & 0-588 \\
                downtown\_weeklyMarket\_00 & 0-1192 \\
                downtown\_bus\_00& 0-800 \\
                office\_phoneCall\_00 & 0-879 \\
				downtown\_walkBridge\_01 & 156-1371 \\
				downtown\_runForBus\_01 & 300-485 \\
				downtown\_sitOnStairs\_00 & 84-918 \\
                downtown\_bar\_00 & 66-256 \\
                downtown\_cafe\_00 & 128-736 \\
                downtown\_enterShop\_00 & 0-1448 \\\
                downtown\_rampAndStairs\_00 & 0-219 \\
                downtown\_runForBus\_00 & 88-350 \\
                downtown\_windowShopping\_00 & 0-1750 \\
                downtown\_crossStreets\_00 & 152-587 \\
                downtown\_car\_00 & 0-576 \\
                downtown\_walking\_00 & 0-1386 \\
				\hline
		\end{tabular}
	\end{center}\vspace{-5mm}
	\caption{Video sequences of Dyna3DPW. }\label{tab:Dyna3DPW}
\end{table}

\textbf{3DPW} is a multi-person in-the-wild video dataset.
Videos in 3DPW are captured by tracking subjects with a dynamic camera to record their daily activities, which matches our task.
3DPW contains rich human activities, such as shopping in market, running for the bus, walking on crowded bridge, and working in an office.
3DPW provide  accurate 3D human pose and mesh annotations of the tracked subjects.
Note that only the one or two subjects are labeled and all other people appearing in the images are not
%Note that pedestrians are not labelled.
Since the global trajectory annotations are not accurate, we evaluate the 3D human pose and shape estimation on the test set of 3DPW.

\textbf{Dyna3DPW} is a subset of 3DPW test set that we constructed. 
Not all 3DPW videos have complete tracking annotations due to occlusion.
Consequently, we select the 16 video sequences in Tab.~\ref{tab:Dyna3DPW} that are complete and call this subset Dyna3DPW.
There are many challenging scenes in Dyna3DPW, such as pedestrians walking past the camera, occlusion between tracked subjects, object occlusion, truncation, and sudden changes of direction.
We use this challenging subset to evaluate tracking and HPS accuracy in complex scenes with a moving camera.

\textbf{Human3.6M~\cite{h36m}} and \textbf{MPI-INF-3DHP}~\cite{mono-3dhp2017} are single-person 3D pose video datasets. 
They are captured in constrained experimental environments using static multi-view cameras. 
They provides 3D pose annotations for each frame.
We sample every 5 frames to reduce redundancy.
We use their training set for training.

We also use two 2D human pose datasets, PennAction~\cite{zhang2013actemes} and PoseTrack~\cite{doering2022posetrack21}. 
\textbf{PennAction} contains 2326 video sequences of 15 different actions.
Most videos in PennAction contain only a single person with 2D pose annotations. 
Part of videos containing multiple people while only a specific subject has 2D pose annotations. 
\textbf{PoseTrack} is an in-the-wild 2D video dataset for multi-person pose tracking. 
It contains 763 video sequences and provides 2D pose and ID annotations.
%However, the video is not captured by following specific subjects and recording their activities. 
We use these for training. 

\section{Experiments with Static Cameras}

\begin{table}[t]
\footnotesize
\begin{center}
\begin{tabular}{cccc}
    \hline
    HMR~\cite{hmr} & CRMH~\cite{jiang2020coherent} & ROMP~\cite{romp} & TRACE \\
    \hline
    56.8 & 52.7 & 50.2 & \textbf{42.0}\\
    \hline
    \end{tabular}
    \end{center}\vspace{-5mm}
\caption{Comparisons on Human3.6M using Protocol 2 of HMR~\cite{hmr} in PAMPJPE.}\label{tab:h36m}
\end{table}

\begin{table}[t]
\footnotesize
\begin{center}
\begin{tabular}{cccc}
    \hline
    CRMH~\cite{jiang2020coherent} & ROMP~\cite{romp} & Pose2UV~\cite{Pose2UV} & TRACE \\
    \hline
    102.6 & 72.0 & 69.5 & \textbf{38.2}\\
    \hline
    \end{tabular}
    \end{center}\vspace{-5mm}
\caption{Comparisons on 3DMPB~\cite{Pose2UV} in PAMPJPE. }\label{tab:3DMPB}
\end{table}

\begin{table}[t]
\footnotesize
\begin{center}
\begin{tabular}{cccc}
    \hline
    ID switches & MOTA & IDF1 & HOTA \\
    \hline
    1 & 96.1 & 98.0 & 64.8\\
    \hline
    \end{tabular}
    \end{center}\vspace{-5mm}
\caption{Tracking results on CMU Panoptic~\cite{cmu_panoptic}. }\label{tab:cmup_tracking}
\end{table}

As a sanity check, we also evaluate the 3D human pose estimation and tracking with static cameras.
On indoor Human3.6M, we follow the evaluation protocol 2 of HMR~\cite{hmr}. 
As results shown in Tab.~\ref{tab:h36m}, compared with previous SOTA methos, TRACE achieves comparable results.
On in-the-wild 3DMPB~\cite{Pose2UV}, we follow Pose2UV~\cite{Pose2UV} to report PAMPJPE in Tab.~\ref{tab:3DMPB} and achieve new SOTA results. 
We evaluate tracking on severe occlusion scenes from CMU Panoptic~\cite{cmu_panoptic}, selected by CRMH~\cite{jiang2020coherent}.
%We report: ID switches (1), MOTA (96.1), IDF1 (98.0), HOTA (64.8).
As shown in Tab.~\ref{tab:cmup_tracking}, TRACE achieves promising results.
These results testify the performance of TRACE with static cameras. 

\textbf{Runtime efficiency.} 
TRACE runs at 32.2 FPS on MuPoTS while using 6.9GB of memory on a 4090 GPU. TRACE takes 174.1G FLOPs, which includes the image backbone, motion backbone, and heads, accounting for 41.0G, 66.4G, and 66.7G, respectively. 

\section{More Ablation Studies}

\textbf{Memory unit.} Without the memory unit, on MuPoTS/Dyna3DPW, ID switches increase 31/94 and MOTA, IDF1, and HOTA decreases 6.5\%/11.1\%, 12.0\%/49.1\%, and 8.0\%/39.6\% respectively.
These results demonstrate that the memory unit is an essential module for persistent tracking under occlusion.
It enables that the predicted 3D motion offset can associate the occluded objects in memory with new detections.

\textbf{Backbones.} To enable end-to-end 5D Representation Learning (5DRL), we develop Temporal Feature Propagation (TFP) as a differentiable path for modeling temporal image features. On 3DPW, 5DRL with TFP only (w/o the optical flow from the motion backbone) outperforms BEV~\cite{bev} by 18.1\% in terms of PAMPJPE (38.4mm).
Simultaneously learning temporal motion features from the motion backbone further reduces PAMPJPE by 1.6\% (37.8mm).

\section{Discussion of Limitations}
 
As ours is the first method to perform one-stage global 3D human motion and trajectory estimation from videos captured by dynamic cameras, TRACE has some limitations that could be further explored in future work.

Due to the lack of training data for this task, we make restrictive assumptions to facilitate effective training. 
For example, we assume all videos are captured by a camera with fixed field of view (FOV=50 degree), without shot changes~\cite{pavlakos2022sitcoms3d}.
Additionally, due to the low diversity of human body shapes in the training datasets, 3D human body shape estimation is limited. 
Collecting or synthesizing a new dataset with sufficient diversity in camera and human body shape could potentially solve these problems. 
We also assume the camera coordinate of the first frame of a input video as the world coordinates. 
Future work should work on estimating the camera poses (e.g.~pitch, yaw, and roll) from the monocular video to convert the results of TRACE to the real world coordinates. 
Here we focused on the problem of tracking specific subjects that are specified in the first frame.
Our method, like BEV~\cite{bev}, actually finds all the people in the image but the tracking branch filters out all but the subjects.
Future work should explore extending TRACE to track everyone in the scene.

{\small
\bibliographystyle{ieee_fullname}
\bibliography{egbib}
}

%% file: definitions.tex
\newcommand{\smpl}{\mbox{SMPL}\xspace}
\newcommand{\hcolor}{black}

%% file: Sections/abstract.tex
%%%%%%%%% ABSTRACT
\begin{abstract}
Although the estimation of 3D human pose and shape (HPS) is rapidly progressing, current methods still cannot reliably estimate \textcolor{\hcolor}{moving} humans in global coordinates, which is critical for many applications. This is particularly challenging when the camera is \textcolor{\hcolor}{also} moving, entangling human and camera motion. To address these issues, we adopt a novel 5D representation (space, time, and identity) that enables end-to-end reasoning about people in scenes. Our method, called TRACE, introduces several novel architectural components. Most importantly, it uses two new ``maps" to reason about the 3D trajectory of people over time in camera, and world, coordinates. An additional memory unit enables persistent tracking of people even during long occlusions. TRACE is the first one-stage method to jointly recover and track 3D humans in global coordinates from dynamic cameras. By training it end-to-end, and using full image information, TRACE achieves state-of-the-art performance on tracking and recovering global human trajectories. The code\footnote{\url{https://www.yusun.work/TRACE/TRACE.html}} and dataset\footnote{\url{https://github.com/Arthur151/DynaCam}} are released for research purposes.\footnote{Please note this is a modified version of the original paper that fixes an error in the evaluation. Please refer to Sec.~\ref{sec:erratum} for details.}

\end{abstract}

%% file: Sections/introduction.tex
%%%%%%%%% BODY TEXT
\section{Introduction}
\label{sec:intro}
The estimation of 3D human pose and shape (HPS) has many applications and there has been significant recent progress \cite{hmr,sun2019dsd-satn,kolotouros2019spin,kocabas2020vibe,zhang2021pymaf,Kocabas_SPEC_2021,moon2020pose2pose,pavlakos2019texturepose,keep,yi2022mime,yi2022generating,tripathi2023ipman,BEDLAM}.
Most methods, however, reason only about a single frame at a time and estimate humans in {\em camera coordinates}.
Moreover, such methods do not {\em track} people and are unable to recover their global trajectories.
The problem is even harder in typical hand-held videos, which are filmed with a dynamic, moving, camera.
For many applications of HPS, single-frame estimates in camera coordinates are not sufficient.
To capture human movement and then transfer it to a new 3D scene, we must have the movement in a coherent global coordinate system.
This is a requirement for computer graphics, sports, video games, and extended reality (XR).

Our key insight is that most methods estimate humans in 3D, whereas the true problem is 5D.
That is, a method needs to reason about 3D space, time, and subject identity.
With a 5D representation, the problem becomes tractable, enabling a {\em holistic} solution that can exploit the full video to infer multiple people in a coherent global coordinate frame.
As illustrated in Fig.~\ref{fig:teaser}, we develop a unified method to jointly regress the 3D pose, shape, identity, and global trajectory of the subjects in global coordinates from monocular videos captured by dynamic cameras (DC-videos).

To achieve this, we deal with two main challenges. 
First, DC-videos contain both human motion and camera motion and these must be disentangled to recover the  human trajectory in global coordinates.
One idea would be to recover the camera motion relative to the rigid scene using structure-from-motion (SfM) methods (e.g.~\cite{liu20214d}).
In scenes containing many people and human motion, however, such methods can be unreliable. 
An alternative approach is taken by GLAMR \cite{yuan2022glamr}, which infers global human trajectories from local 3D human poses, without taking into account the full scene.
By ignoring evidence from %about the scene that is present in
the full image, GLAMR fails to capture the correct global motion in common scenarios, such as biking, skating, boating, running on a treadmill, etc.
Moreover, GLAMR is a multi-stage method, with each stage dependent on accurate estimates from the preceding one.
Such approaches are more brittle than our holistic, end-to-end, method.

\iffalse
in DC-videos, as human motions are coupled with high-frequency camera movement (caused by shaking, running, etc), it is hard to estimate the  human trajectory in global coordinates.
A few methods~\cite{liu20214d,yuan2022glamr} have made some attempts on solving this challenge from some specific image clues.
For instance, Liu et. al.~\cite{liu20214d} have employed a classical structure-from-motion (SFM) method~\cite{schonberger2016structure} to estimate camera poses from DC-videos via detecting cross-frame correspondences (e.g. environment keypoints).
However, since our input video is uncalibrated and moving objects occupy a large portion of the image, it is difficult for SFM methods to distinguish the difference between camera rotation and translation under such highly under-determined conditions.
GLAMR~\cite{yuan2022glamr} bypasses the ambiguity of monocular camera poses and infers global human trajectories from local 3D human poses only.
However, ignoring environmental information will lead to failures in many scenarios, such as biking, skating, boating, running on a treadmill, etc.
\fi 

The other challenge, as shown in the upper right corner of Fig.~\ref{fig:teaser}, is that severe occlusions are common in videos with multiple people.
Currently, the most popular tracking strategy is to infer the association between 2D detections using a temporal prior (e.g.~Kalman filter)~\cite{zhang2022bytetrack}. 
However, in DC-videos, human motions are often irregular and can easily violate hand-crafted priors.
PHALP \cite{rajasegaran2022tracking} is one of the few methods to address this for 3D HPS.
It uses a classical, multi-stage, detection-and-tracking formulation with heuristic temporal priors.
%that builds an appearance model of the people being trackt and uses this to 
It does not holistically reason about the sequence and is not trained end-to-end.

To address these issues, we reason about people using a 5D representation and capture information from the full image and the motion of the scene.
This holistic reasoning enables the reliable recovery of global human trajectories and subject tracking using a single-shot method.
This is more reliable than multi-stage methods because the network can exploit more information to solve the task and is trained end-to-end.
No hand-crafted priors are needed and the network is able to share information among modules.

\begin{figure*}[t]
	\centerline{\includegraphics[width=1.00\textwidth]{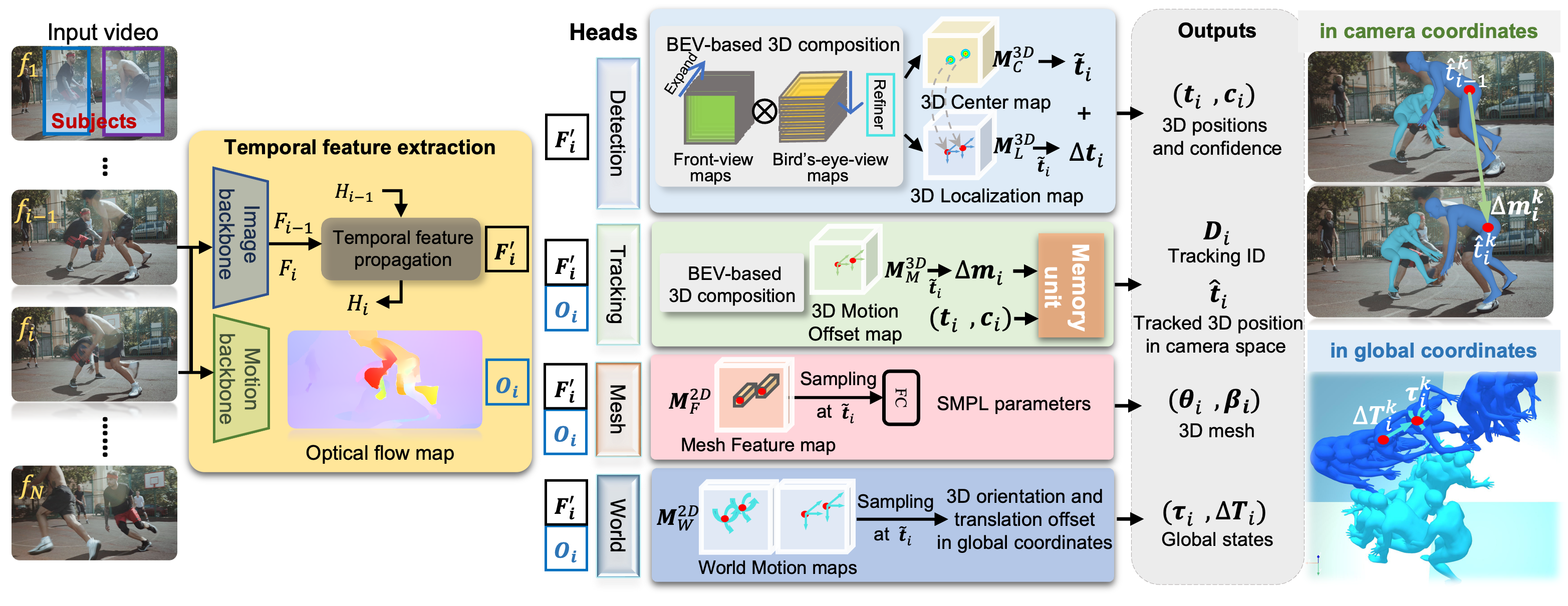}}
	\vspace{-3mm}
	\caption{{\bf TRACE Overview}. TRACE takes a video sequence and regions in the first frame corresponding to the subjects to be tracked. TRACE encodes the video and its motion with temporal features and optical flow. A novel 3D Motion Offset map reasons about human trajectories in camera coordinates. The World Motion map represents the trajectory in global coordinates. A memory unit deals with occlusions by encoding the subject identities. TRACE is trained end-to-end to estimate the 3D  shape and pose of multiple people throughout a video in global coordinates.
 See Sec.~\ref{sec:overview} for details.}\vspace{-2mm}
	\label{fig:framework}
\end{figure*}

Specifically, we develop TRACE, a unified one-stage method for \textbf{T}emporal \textbf{R}egression of  \textbf{A}vatars with dynamic \textbf{C}ameras in 3D \textbf{E}nvironments. 
The architecture is inspired by BEV~\cite{bev}, which directly estimates multiple people in depth from a single image using multiple 2D maps.
BEV uses a 2D map representing an imaginary, ``top down", view of the scene. 
This is combined with an image-centric 2D map to reason about people in 3D.
Our key insight is that the idea of maps can be extended to represent how people {\em move in 3D}.
With this idea, TRACE  %takes the idea of maps representing the bodies in the scene and 
introduces three new modules to holistically model 5D human states, performing multi-person temporal association, and inferring human trajectories in global coordinates; see Fig.~\ref{fig:framework}.

First, to construct a holistic 5D representation of the video, we extract temporal image features by fusing single-frame feature maps from the image backbone with a temporal feature propagation module.
We also compute the optical flow between adjacent frames with a motion backbone.
The optical flow provides short-term motion features that carry information about the motion of the scene and the people.
%to further enhance the short-term motion features. 
Second, to explicitly track human motions, we introduce a novel {\em 3D motion offset map} to establish the association of the same person across adjacent frames.
This map contains a 3D offset vector at each position, which represents the difference between the 3D positions of the same subject from the previous frame to the current frame in camera coordinates. 
We also introduce a {\em  memory unit} to keep track of subjects under long-term occlusion.
\textcolor{\hcolor}{Note that the 3D trajectories are built in camera space, and TRACE uses a novel {\em world motion map} that transfers the trajectories to global coordinates.}
%To explicitly learn the global trajectory of people in world coordinates, we introduce a {\em world motion map}.
At each position, this map contains a 6D vector to represent the difference between the 3D positions of the corresponding subject from the previous frame to the current frame and its 3D orientation in world coordinates. 
Taken together, this novel network architecture goes beyond prior work by taking information from the full video frames to address detection, pose estimation, tracking, and occlusion in a holistic network that is trained end-to-end.

To enable training and evaluation of global human trajectory estimation from in-the-wild DC-videos, we build a new dataset, DynaCam.
Since collecting global human trajectories and camera poses with in-the-wild DC-videos is difficult,  we simulate a moving camera using publicly available in-the-wild panoramic videos and regular videos captured by static cameras.
In this way, we create more than 500 in-the-wild DC-videos with precise camera pose annotations.
Then we generate pseudo-ground-truth 3D human annotations via fitting~\cite{joo2020eft} SMPL~\cite{smpl} to detected 2D pose sequences~\cite{ge2021yolox,zhang2022bytetrack,xu2022vitpose}.
With 2D/3D human pose and camera pose annotations, we can obtain the global human trajectories using the PnP algorithm~\cite{fischler1981random}. 
This dataset is sufficient to train TRACE to deal with dynamic cameras. %multiple people and dynamic cameras.

We evaluate TRACE on a multi-person in-the-wild benchmark (MuPoTS-3D~\cite{mehta2018single}) and our DynaCam dataset. 
%On 3DPW, TRACE achieves the state-of-the-art (SOTA) PA-MPJPE of 37.8mm, less the current best (42.7 \cite{li2022dnd}).
On MuPoTS-3D, TRACE outperforms previous 3D-representation-based methods~\cite{rajasegaran2021tracking,rajasegaran2022tracking} and tracking-by-detection methods~\cite{zhang2022bytetrack} on tracking people under long-term occlusion. 
On DynaCam, TRACE outperforms GLAMR~\cite{yuan2022glamr} in estimating the 3D human trajectory in global coordinates from DC-videos. 

In summary, our main contributions are: 
(1) We introduce a 5D representation and use it to learn holistic temporal cues related to both 3D human motions and the scene.
(2) We introduce two novel motion offset representations to explicitly model temporal multi-subject association and global human trajectories from temporal clues in an end-to-end manner.
(3) We estimate long-term 3D human motions over time in global coordinates, achieving SOTA results.
(4) We collect the DynaCam dataset of DC-videos with pseudo ground truth, which facilitates the training and evaluation of global human trajectory estimation.
\textcolor{\hcolor}{The code and dataset are publicly available for research purposes.}

%% file: Sections/related_work.tex
\section{Related Work}
\label{sec:related_work}
\textbf{Monocular 3D mesh regression with full images.}
Most existing methods \cite{hmr,sun2019dsd-satn,kolotouros2019spin,kocabas2020vibe,zhang2021pymaf,Kocabas_SPEC_2021,moon2020pose2pose,pavlakos2019texturepose,li2022dnd,zanfir2018monocular,xiu2022icon,xiu2023econ,liao2023car} take a multi-stage detection-based pipeline to estimate 3D HPS from  cropped image patches, which exclude important cues, such as camera information and human-scene relationships.
A few recent multi-stage~\cite{li2022cliff} and one-stage~\cite{romp,bev} methods have made steps towards using the full-image information. 
For instance, CLIFF~\cite{li2022cliff} estimates 3D HPS  
\textcolor{\hcolor}{by taking into account the bounding box locations, giving the method camera information and improving accuracy.}
To directly estimates multiple people at once from the full image, ROMP~\cite{romp} introduces a 2D Center heatmap and a Mesh parameter map to represent 2D human locations and 3D human body meshes, respectively. 
%ROMP achieves this using a 2D Center heatmap and a Mesh parameter map to represent 2D human locations and 3D human body meshes, respectively. 
BEV~\cite{bev} goes beyond ROMP by introducing an imaginary bird's-eye-view map, which is combined with the front-view maps to construct a 3D view in camera coordinates. 
However, they only model 3D HPS in camera coordinates from a single image. 
%However, all these methods only model in a single image frame and represent 3D humans in camera coordinates.
%Like ROMP and BEV, TRACE looks at the whole image and uses the idea of ``maps".  
Using ``maps" like ROMP and BEV, TRACE also looks at the full image. 
%We go further, however, by introducing novel maps that model the coherent motion of multiple people across a video sequence in world coordinates.
We go further, however, by introducing novel maps that model human motions across a video sequence in global coordinates.
%Here, we  a more holistic view of the video sequence. 

\textbf{Tracking datasets.}
While there are many tracking datasets~\cite{dendorfer2020mot20,dave2020tao,Geiger2012CVPR,Shanshan2017CVPR,ess2008mobile,caelles20192019} with 2D annotations,
% (e.g.~2D bounding boxes and 2D segmentations).
%, such as MOTChallenge~\cite{leal2015motchallenge,milan2016mot16,dendorfer2020mot20}, TAO~\cite{dave2020tao}, KITTI~\cite{Geiger2012CVPR}, CityPersons~\cite{Shanshan2017CVPR}, ETHZ~\cite{ess2008mobile}, and DAVIS~\cite{caelles20192019}. 
only a few~\cite{fabbri2021motsynth,ehsanpour2022jrdb} capture the 3D trajectory of pedestrians.
In both cases, the scene and human activities are limited. %, like MOTSynth~\cite{fabbri2021motsynth} and JRDB-Act~\cite{ehsanpour2022jrdb},
%\textcolor{red}{Here we use MuPoTS (cite and explain?) and 3DPW (cite and explain?).}
To address this, we use 3DPW~\cite{3dpw} and MuPoTS-3D~\cite{mehta2018single} for tracking evaluation.
3DPW is the most relevant dataset for our task and it provides a real-world test case.
3DPW contains videos that are captured by a moving camera that follows subjects to record their activities in many daily scenes. %, having coffee, parkour, going to a bar, etc. going up and down stairs, , such as shopping, running for a bus, chatting, etc
%and its setup perfectly matches our task. 
%The evaluation on 3DPW can be regarded as a practical use test.
MuPoTS-3D contains rich multi-person interaction scenes with long-term occlusions for tracking evaluation. 
%and the difficulty of direct world trajectory estimation, tracking in camera coordinates is more reliable than in world trajectory.
%Although many tracking datasets~\cite{leal2015motchallenge,milan2016mot16,dendorfer2020mot20,dave2020tao,Geiger2012CVPR,Shanshan2017CVPR} provide 2D annotations (e.g.~ bounding boxes and segmentation) and some 3D datasets lable the 3D trajectories of pedestrians~\cite{fabbri2021motsynth, ehsanpour2022jrdb}, the scale and diversity of scenes and human activities of them are limited.

\textbf{Tracking 3D people through occlusions.}
Most existing methods~\cite{bergmann2019tracking,zhang2022bytetrack,ruan2019poinet,xiu2018poseflow} perform tracking using 2D image cues.
%, such as bounding box detections~\cite{bergmann2019tracking,zhang2022bytetrack}, 2D keypoints~\cite{ruan2019poinet,xiu2018poseflow}, and 2D segmentations~\cite{Wang_2019_CVPR}.
%For instance, following a classic % optical flow~\cite{xiao2018simple}, %, and 2D appearance~\cite{meinhardt2022trackformer}. 
The classic tracking-by-detection paradigm focuses on associating the 2D detections using a temporal prior (e.g.~Kalman filter).
When applied to DC-videos containing rapid human and camera motions that violate the hand-crafted priors, such methods are brittle.
%They show impressive performance with static cameras, but do not exploit temporal image features.
%This makes them brittle when applied to DC-videos containing rapid human and camera motions that violate their hand-crafted prior.
%Additionally, 2D information is ambiguous under severe occlusions. 
%To alleviate the ambiguity in 2D,
Going beyond 2D,
PHALP~\cite{rajasegaran2022tracking} separately extracts 3D human pose, appearance, and location with a multi-stage design from each video frame, and then assembles them for tracking.
%Such multi-stage frameworks are always susceptible to errors in early stages propagating to later ones.
%While errors accumulate in these multi-stage frameworks and under long-term occlusions all these 3D cues are invisible.
In contrast to these multi-stage methods, which are susceptible to errors in early stages,
we explicitly learn the 3D human trajectory from temporal 5D cues in an end-to-end manner.
%We also develop a memory unit to perform persistent tracking under long-term occlusions. 
%\textcolor{\hcolor}{
%Different from most methods stop at tracking in camera coordinates, we further develop novel world maps to transfer the trajectories to global coordinates.
%Since there is more training data in camera coordinates, than in world coordinates, learning to track humans in camera coordinates would be much easier. But most methods stop there. We develop novel world maps to transfer the trajectories to global coordinates.}

\textbf{Monocular global 3D human trajectory reasoning.}
Most existing methods~\cite{zheng2021deepmulticap} that reason about global 3D human trajectories do so with static, calibrated, cameras in a multi-view setting.
A few recent methods~\cite{liu20214d,yuan2022glamr} have addressed the ill-posed problem of extracting the global motions of humans from monocular video.
Liu et al.~\cite{liu20214d} employ a structure-from-motion (SfM) method~\cite{schonberger2016structure} to estimate the camera poses from monocular videos captured by a dynamic camera. 
However, when the input video contains the movement of multiple subjects, it is hard for SfM methods to extract sufficiently many stable keypoints for reliable camera estimation.
%Besides, it is hard to distinguish the 3D translation and rotation movement of dynamic camera from limited visual differences between adjacent frames. 
GLAMR~\cite{yuan2022glamr} adopts a multi-stage pipeline to infer the global human trajectory from root-relative local human 3D poses estimated from each frame.
\textcolor{\hcolor}{The per-frame human pose estimates make it vulnerable to occlusion.}
%The quality of their global trajectory reasoning is highly dependent on the quality of per-frame 3D human pose estimation, which makes it vulnerable to occlusions (either person-person occlusion or truncation, such as running with legs occluded).
Additionally, GLAMR relies on bounding boxes, ignoring scene-related information.
Consequently, GLAMR fails in common scenarios like riding a bike or skating. % (e.g.~riding and skating).
\textcolor{\hcolor}{In concurrent work (in this proceedings), SLAHMR \cite{SLAHMR} uses a multi-stage optimization-based approach that combines structure from motion with human motion priors to estimate 4D human trajectories in global coordinates; this is very computationally expensive. 
In contrast to previous multi-stage methods, TRACE simultaneously combines scene information and 3D human motions with a novel 5D representation to holistically exploit all temporal cues and to enable end-to-end training.
%The 5D representation is the key that enables our approach to be trained end-to-end.
}

%% file: Sections/method.tex
\section{Method}\vspace{-1mm}
\label{sec:method}

\subsection{Overview}\vspace{-1mm}
\label{sec:overview}

The overall framework of TRACE is shown in Fig.~\ref{fig:framework}.
Given a video sequence captured with a dynamic camera $\{f_i,i=1,\dots,N\}$ with $N$ frames, the user specifies $K$ tracking subjects shown in the first frame. Our goal is to simultaneously recover the 3D pose, shape, identity, and  trajectory of each subject in global coordinates.
To achieve this, TRACE first extracts temporal features and then decodes each sub-task with a separate head network. 
First, via two parallel backbones, TRACE encodes the video and its motion into temporal image feature maps $\boldsymbol{F}^{\prime}_i$ and motion feature maps $\boldsymbol{O}_{i}$.
%Specifically, for subject $k$, we reconstruct the global motion ${\boldsymbol{G}^k}$, including the 3D pose parameter $\boldsymbol{\Theta}^k=(\boldsymbol{\theta}_1^k,...,\boldsymbol{\theta}_N^k)$, 3D shape parameter $\boldsymbol{B}^k=(\boldsymbol{\beta}_1^k, ..., \boldsymbol{\beta}_N^k)$, global trajectory $\boldsymbol{\Theta}^k$ in $N$ frames are human mesh parameters in SMPL~\cite{smpl} model. With the global trajectory $\boldsymbol{T}^k$, we reconstruction the 3D motions in the world coordinates.

The Detection and Tracking branches take these features and perform multi-subject tracking in {\em camera coordinates}.
%while the tracking branch is completely new.
%We then estimate in camera coordinates for 3D detection and multi-subject tracking.
Unlike BEV~\cite{bev}, our detection method takes {\em temporal} image features $\boldsymbol{F}^{\prime}_i$ as input.
It uses the features to detect the 3D human positions $\boldsymbol{t}_i$ and their confidence $\boldsymbol{c}_i$ for all people in frame $f_i$. 
The Mesh branch regresses all the human mesh parameters $(\boldsymbol{\theta}_i,\boldsymbol{\beta}_i)$, in SMPL~\cite{smpl} format, from the input Feature maps.
Unlike BEV, this branch takes both temporal image features and  motion features.

The combined features ($\boldsymbol{F}^{\prime}_i$,$\boldsymbol{O}_{i}$) are fed to our novel Tracking branch to estimate the 3D Motion Offset map, indicating the 3D position change of each subject across frames. 
The new Memory Unit takes the 3D detection and its 3D motion offset as input.
It then determines the subject identities and builds human trajectories $\boldsymbol{\widehat{t}}_{i}$ of the  $K$ subjects in camera coordinates. 
Note that, like BEV, our detection branch finds all the people in the video frames but our goal is to track only the $K$ input subjects.
Consequently, the memory unit filters out detected people who do not match the subject trajectories.
%Note that non-subject detections are filtered out in this step.

Finally, to estimate subject trajectories in global coordinates, the World branch 
estimates a world motion map, representing the 3D orientation $\boldsymbol{\tau}_i$ and 3D translation offset $\Delta \boldsymbol{T}_i$ of the $K$ subjects in global coordinates. 
Accumulating $\Delta \boldsymbol{T}_i$, starting with the 3D position $\boldsymbol{\widehat{t}}_{1}$ of the tracked subjects in the first frame, gives their global 3D trajectory $\boldsymbol{T}$.
\textcolor{\hcolor}{Note that the global (``world") coordinates are defined relative to the camera coordinates of the first frame.}
%we  obtain their global 3D trajectory $\boldsymbol{T}=\{\boldsymbol{\widehat{t}}_{1}, \boldsymbol{\widehat{t}}_{1}+{\Delta \boldsymbol{T}_2}, \boldsymbol{\widehat{t}}_{1}+{\Delta \boldsymbol{T}_2}+{\Delta \boldsymbol{T}_3},\dots\}$ spanning the video.
%Combining these results, we obtain the 3D motion and trajectory of the $K$ subjects in both camera and world coordinates.
%Note that we take the camera coordinates in the first frame as the world coordinates for the video; everything is then relative to those.
%The initial world coordinates of the subjects are then their initial camera coordinates.
%The world coordinates can converted to any a specific 3D scene via calibrating the first frame coordinates.
%The whole network is trained end-to-end to estimate the global motion ${\boldsymbol{G}}={(\boldsymbol{\Theta},\boldsymbol{B}, \boldsymbol{T})}$ of multiple people throughout a video.

\subsection{Holistic 5D Representation: Details}
%Our task is to perceive 5D motion state of subjects from video captured by a dynamic camera, including 3D human pose and shape, identity, and global trajectory in world coordinates. 
%A few existing methods~\cite{liu20214d,yuan2022glamr} have made some attempts in directly estimating camera poses from environment keypoints or inferring global human trajectories from local body poses, which leads to limited performance.
%Unlike them using only part of information, we intend to construct a 5D representation with a holistic view of all temporal cues.
Rather than directly estimating camera poses from environment keypoints~\cite{liu20214d} or inferring global human trajectories from local body poses~\cite{yuan2022glamr}, we develop a 5D representation to directly reason about human states, perform multi-person temporal association, and infer human trajectories in global coordinates. 
Learning a holistic 5D representation is the foundation of our one-stage framework.
The representation has five main parts.
%To achieve this, we develop a one-stage framework to learn from all sub-tasks, including human detection, 3D mesh recovery, tracking, and global trajectory estimation. 
%In this section, we introduce how to extract the temporal feature maps and the way of decoding temporal feature maps to the result of sub-tasks.  
%In this sub-section, we introduce how to build the 5D representation, which consists of five main parts.

\textbf{i.~Temporal feature maps.}
To construct the temporal feature maps encoding 5D human states and scenes information, we need to extract both single-frame image features and the motion features between adjacent frames. 
Therefore, given frame $f_{i-1}$ and $f_i$, we adopt a parallel two-branch structure to extract temporal image feature maps $\boldsymbol{F}^{\prime}_i$ and motion feature maps $\boldsymbol{O}_{i}$ for the current frame $f_i$.
%Especially, the image branch is served as a differentiable path to learn more temporal cues in an end-to-end manner. 
First, in the image branch, we extract single-frame feature maps $\boldsymbol{F}_{i-1}$ and $\boldsymbol{F}_i$ with an image backbone (HRNet-32~\cite{cheng2020higherhrnet}).
To extract long-term and short-term motion features, we construct a temporal feature propagation module by combining a ConvGRU~\cite{teed2020raft} module, a Deformable convolution~\cite{zhu2019deformable} module, and a residual connection.
With these, we fuse the image feature maps to generate a temporal image feature map $\boldsymbol{F}^{\prime}_i$.
See Sup.~Mat.~for details and experimental analysis.
Additionally, in the motion branch, we estimate the optical flow map $\boldsymbol{O}_{i}$ between frames $f_{i-1}$ and $f_i$ with a motion backbone (RAFT~\cite{teed2020raft}), to extract motion features of both people and scenes.
From the combined temporal feature ($\boldsymbol{F}^{\prime}_i$, $\boldsymbol{O}_{i}$), then we estimate five maps for the task. 

\textbf{ii.~3D detection maps.} From the temporal image feature $\boldsymbol{F}^{\prime}_i$, we estimate a 3D Center map $\boldsymbol{M}^{3D}_{C} \in \mathbb{R}^{1\times D \times H \times W}$ for coarse human detection and a 3D Localization map $\boldsymbol{M}^{3D}_{L} \in  \mathbb{R}^{1\times D \times H \times W}$ for fine localization. %  via compositing the estimated 2D front-view maps and the 2D bird's-eye-view maps following BEV~\cite{bev}. The 3D detection maps realize reasoning about people in 3D from a 2D image.
The two 3D detection maps are composited from the front-view maps and the 2D bird's-eye-view maps following BEV~\cite{bev}.
%, which reasons about people in 3D from a 2D image.
For $K$ subjects, we first parse out the detected 3D center positions $\boldsymbol{\widetilde{t}}_{i}$ and their detection confidences $\boldsymbol{c}_{i}$ from the 3D Center map. % $\boldsymbol{\widetilde{t_{i}}}$ contains the 3D locations of all the candidate subjects for tracking~~~
Then we sample the 3D Localization map at $\boldsymbol{\widetilde{t}}_{i}$ to obtain the fine 3D localization offset vectors ${\Delta \boldsymbol{t}_{i}}$.
The predicted 3D translation of subjects to be tracked in camera coordinates is $\boldsymbol{t}_{i}=\boldsymbol{\widetilde{t}_{i}}+{\Delta \boldsymbol{t}_{i}}$.

\textbf{iii.~3D Motion Offset map.} To track subjects between adjacent frames using $\boldsymbol{F}^{\prime}_i$ and $\boldsymbol{O}_{i}$, we estimate a 3D Motion Offset map $\boldsymbol{M}^{3D}_{M} \in \mathbb{R}^{3\times D \times H \times W}$ via a BEV-based 3D composition. 
We sample the $\boldsymbol{M}^{3D}_{M}$ at detected 3D center positions $\boldsymbol{\widetilde{t}}_{i}$ to obtain the 3D motion offset vectors ${\Delta \boldsymbol{m}_{i}}$, which represent the 3D position changes of the subjects from frame $f_{i-1}$ to frame $f_{i}$ in camera coordinates.
With ${\Delta \boldsymbol{m}_{i}},\boldsymbol{t}_{i},\boldsymbol{c}_{i}$ as input, a memory unit (Sec.~\ref{sec:memory_unit}) predicts the tracked identities $\boldsymbol{D}_{i}$ and the optimized 3D position $\boldsymbol{\hat{t}}_{i}$ in camera coordinates, which filters out low-confidence detections while keeping track of subjects through long-term occlusion. %. The memory unit

\textbf{iv.~Mesh Feature map.} To regress the \smpl parameters of subjects, we first estimate a 2D Mesh Feature map $\boldsymbol{M}^{2D}_{F} \in \mathbb{R}^{C \times H \times W}$ from temporal features ($\boldsymbol{F}^{\prime}_i$, $\boldsymbol{O}_{i}$).
With the tracked 3D positions $\boldsymbol{\widehat{t}}_{i}=(\boldsymbol{u}_i,\boldsymbol{v}_i,\boldsymbol{d}_i)$, we sample a mesh feature vector from $\boldsymbol{M}^{2D}_{F}$ at 2D positions $(\boldsymbol{u}_i,\boldsymbol{v}_i)$.
To differentiate the features of people at different depths, we map the $\boldsymbol{d}_i$ to a 128-dim encoding vector via an embedding layer and add this with the mesh feature vector to regress SMPL~\cite{smpl} parameters $(\boldsymbol{\theta}_i,\boldsymbol{\beta}_i)$. 
The SMPL model maps the pose and shape, $\boldsymbol{\theta}_i$ and $\boldsymbol{\beta}_i$, to a 3D human body mesh $\boldsymbol{B} \in \mathbb{R}^{6890 \times 3}$.
With a sparse weight matrix $\boldsymbol{R} \in \mathbb{R}^{Q \times 6890}$ that describes the mapping from $\boldsymbol{B}$ to $Q$ body keypoints, we can obtain the 3D positions of body keypoints $\boldsymbol{J} \in \mathbb{R}^{Q \times 3}$ via $\boldsymbol{R}\boldsymbol{B}$.
With the estimated 3D positions $\boldsymbol{\widehat{t}}_{i}$ and a pre-defined camera projection matrix $\boldsymbol{P}$, the projected 2D keypoints are $\boldsymbol{J}_{2D}=\boldsymbol{P}(\boldsymbol{J}+\boldsymbol{\widehat{t}}_{i})$.

% here 
\textbf{v.~World Motion map.} $\boldsymbol{M}^{2D}_{W} \in \mathbb{R}^{6 \times H \times W}$ contains the 6D vectors that describe the 3D orientation and the relative translation of the subjects in global coordinates. 
We sample $\boldsymbol{M}^{2D}_{W}$ at the 2D positions $(\boldsymbol{u}_i,\boldsymbol{v}_i) \in \boldsymbol{\widehat{t}_{i}}$ of the tracked subjects to obtain their 3D body orientation $\boldsymbol{\tau}_i$ and 3D body motion offset $\Delta \boldsymbol{T}_i$. For the tracked subject $k$,
$\Delta \boldsymbol{T}_i^k$ represents their 3D position change from frame $f_{i-1}$ to frame $f_i$ in global coordinates.
We take the camera coordinate system of the first frame as the global coordinate system for the video.
%\textcolor{\hcolor}{note that to put our output into some other scene with a different world coordinate system,}
The global 3D trajectories of all tracked subjects are obtained by 
accumulating $\Delta \boldsymbol{T}_i$ to their 3D position $\boldsymbol{\widehat{t}}_{1}$ in the first frame,  $\boldsymbol{T}=\{\boldsymbol{\widehat{t}}_{1}, \boldsymbol{\widehat{t}}_{1}+{\Delta \boldsymbol{T}_2}, \boldsymbol{\widehat{t}}_{1}+{\Delta \boldsymbol{T}_2}+{\Delta \boldsymbol{T}_3},\dots\}$.
Combining the 3D mesh, ID, and the global trajectories,  the network reasons about 3D motions and trajectories of $K$ tracked subjects in global coordinates.

%The world coordinates of any a specific 3D scene can be converted via calibrating the first frame coordinates
% why it can reason World Coor? not very clear 

%$\Delta \boldsymbol{T}_i$ represents its 3D position change of the subject from the frame $f_{i-1}$ to frame $f_i$ in world coordinates.

%For recovering the global trajectory of subjects in world coordinates, we estimate the World Motion maps $\boldsymbol{M}^{2D}_{W} \in \mathbb{R}^{6 \times H \times W}$, representing the 3D translation and 3D orientation of subjects in world coordinates. We sample the $\boldsymbol{M}^{2D}_{W}$ at 2D position $(\boldsymbol{u}_i,\boldsymbol{v}_i) \in \boldsymbol{\widehat{t}_{i}}$ of tracked subjects to obtain their 3D body orientation $\boldsymbol{\tau}_i$ and 3D body motion offset $\Delta \boldsymbol{T}_i$ in world coordinates. $\Delta \boldsymbol{T}_i$ represents the 3D position change of the subject from the frame $f_{i-1}$ to frame $f_i$ in world coordinates.

\subsection{Tracking with a Memory Unit}
\label{sec:memory_unit}
We construct the 3D trajectory of each subject by associating the 3D detections $\boldsymbol{t}_{i}$ over time with a 3D motion offset $\Delta \boldsymbol{m}_i$.
To deal with long-term occlusions, we design the Memory Unit for persistent tracking, which will keep the memory for the full sequence.
%Exsiting methods (e.g. ByteTrack~\cite{zhang2022bytetrack}) employ temporal priors (e.g. Kalman Filter) to infer the future position of each subject to associate with the new detections. 
%However, since our input videos are captured by dynamic cameras, human motions in the frames are often irregular, which can easily break such hand-crafted priors.
%Unlike previous methods~\cite{zhang2022bytetrack,bergmann2019tracking} using , we explicitly estimate the temporal human motion with a 3D motion offset $\Delta \boldsymbol{m}_i$ for association from temporal cues in an end-to-end manner.
The memory unit stores the human states during inference and is not used for training. 
With predicted 3D positions $\boldsymbol{t}_{i}$, detection confidences $\boldsymbol{c}_i$, and 3D motion offsets $\Delta \boldsymbol{m}_i$ as inputs, the memory unit can track online.
In each process, we have three stages.
%initialization, memory node matching, and memory update. 

\textbf{i.~Initialization.}
First, we discard predicted 3D positions $\boldsymbol{t}_{i}=(\boldsymbol{x}_i,\boldsymbol{y}_i,\boldsymbol{z}_i)$
%We filter out the candidates in $\boldsymbol{t}_{i}$ 
whose detection confidence is below a threshold $\lambda_c$.
%Considering that our input video is usually shot by tracking the subjects, 
We observe that our input video is usually shot by tracking the subjects, therefore, we discard the detection whose $1/\boldsymbol{z}_i$ is below the scale threshold $\lambda_s$.
To suppress duplicate detections, we find the detection pairs whose Euclidean distance is below a pre-defined threshold $\lambda_d$ and discard the detection with lower detection confidence.
In the first frame, we use the 3D positions $\boldsymbol{t}_{1}$and detection confidences $\boldsymbol{c}_i$ of $K$ subjects to initialize the $K$ memory nodes.

\textbf{ii.~Memory node matching.}
The memory nodes store the 3D position $\boldsymbol{t}^k_{i-1}$ of subject $k$ in previous frames.
We match the memory node with the new filtered detections in the current frame.
We select the predicted 3D translation $\boldsymbol{t}_{i}$ that best matches the 3D trajectory as the optimized 3D position $\boldsymbol{\hat{t}}_{i}$.
Specifically, with filtered 3D positions $\boldsymbol{t}_{i}$ and 3D motion offset $\Delta \boldsymbol{m}_i$, we calculate the Euclidean distance matrix between $\boldsymbol{W}\boldsymbol{t}^k_{i-1}$ and $\boldsymbol{W}(\boldsymbol{t}_{i}-\Delta \boldsymbol{m}_i)$ 
%for Hungarian matching 
where  $\boldsymbol{W}\in\mathbb{R}^{3}$ is a distance weight vector.
To avoid the effects of the non-linear relationship between depth $\boldsymbol{z}_i$ and human scale in images, we convert the depth value $z$ to $1/(1+z)$ when calculating the distance matrix. 
Using Hungarian matching, we keep the matched pairs whose matching distance is below a threshold $\lambda_m$ and use them for memory update.
We use ground truth tracks for training and only perform matching during inference. 

\textbf{iii.~Memory update.}
We update the successfully matched memory nodes with the new 3D position and detection confidence.
For the memory nodes without a matched detection, we accumulate the time since failure. 
Then we remove the memory nodes whose failure time is above a threshold $\lambda_f$.
Tracking can be done in two modes: on-line or off-line. The former does not allow looking back in time, while the latter does.
In the off-line mode, if a new detection re-activates a non-matched memory node, the non-matched part of the 3D trajectory is linearly interpolated.
Finally, the memory unit outputs the latest 3D positions $\boldsymbol{\widehat{t}_{i}}$ and tracking IDs $\boldsymbol{D}_i$ of all memory nodes.

\subsection{DynaCam Dataset}
Even with a powerful 5D representation, we still lack in-the-wild data for training and evaluation of global human trajectory estimation. 
However, collecting global human trajectory and camera poses for natural DC-videos is difficult.
Therefore, we create a new dataset, DynaCam, by simulating camera motions to convert in-the-wild videos captured by static cameras to DC-videos.

We use over 1000 video clips captured by static regular cameras from the MPII Human Pose Database~\cite{andriluka20142d} as well as videos from the InterNet~\cite{pexels}.
\textcolor{\hcolor}{We also use over 200 panoramic video clips that are either recorded by us with an Insta360 RS panoramic camera or are downloaded from the InterNet~\cite{bilibili}.}
%We simulate DC-videos via manual design the 3D translation and rotation of dynamic cameras.
We manually design the 3D rotation and field of view (FOV) of dynamic cameras to track the subjects in panoramic videos.
With the designed camera motions, we can project the panoramic frames into perspective views. 
Also, to simulate the 3D translation of dynamic cameras, we crop the videos captured by static cameras with sliding windows. 
%Please note that cropping-based simulation is an approximation to the real perspective imaging, but it is a cost-effective way to enable an end-to-end solution.
In this way, we can obtain abundant in-the-wild DC-videos with accurate camera pose annotations.
Then we perform 2D human detection, tracking, and 2D pose estimation via YOLOX~\cite{ge2021yolox}, ByteTrack~\cite{zhang2022bytetrack}, and ViTPose~\cite{xu2022vitpose}, respectively, to obtain 2D pose sequences of each subject. We estimate SMPL parameters by fitting the 2D poses using EFT~\cite{joo2020eft} or ProHMR~\cite{kolotouros2021probabilistic} and solve for their 3D positions in camera coordinates via the PnP algorithm (RANSAC~\cite{fischler1981random}). 
Finally, we solve for the 3D human trajectories in the global coordinates with camera pose annotations.
We manually filter out the failure cases.
In this way, we generate more than 500 annotated DC-videos containing over 48K frames.
More than half of video frames are generated from panoramic videos.

\textcolor{\hcolor}{\textbf{Limitations:} 
The videos generated with our process only approximate real DC-videos shot in the wild since they lack perspective effects.
Despite this they prove useful for training TRACE.
%While useful for training TRACE, future work should explore the use of 
%to simulate dynamic camera motion may exhibit slight differences from real-world videos.
} 

\subsection{Loss Functions}
TRACE is supervised by the weighted sum of 15 loss terms that fall into two groups: temporal motion losses and standard image losses.
Here we focus on the novel temporal losses. Please refer to the Sup.~Mat.~for details of all losses.
%The standard image losses are described in Sup.~Mat.

%\textbf{Temporal motion functions.}
%First, we introduce two temporal motion loss terms: 
To learn the temporal motion, we introduce a
3D motion offset loss  $\boldsymbol{\mathcal{L}}_{m}$ and a 6D world motion loss $\boldsymbol{\mathcal{L}}_{W}$.
$\boldsymbol{\mathcal{L}}_{m}$ is the $L_2$ loss between the predicted 3D motion offset ${\Delta \boldsymbol{m}_{i}}$ and ($\boldsymbol{t}^{\prime}_i-\boldsymbol{t}^{\prime}_{i-1}$) where $\boldsymbol{t}^{\prime}_i$ is the ground truth 3D human position at frame $f_i$ in our pre-defined camera coordinates (FOV=$50^{\circ}$), which is solved for via the PnP algorithm~\cite{fischler1981random}.
$\boldsymbol{\mathcal{L}}_{W}$ consists of six parts, including an $L_2$ loss on the global 3D trajectory $\boldsymbol{T}$, an $L_2$ loss on the velocity/acceleration of 3D trajectory nodes $\boldsymbol{\dot{T}}/\boldsymbol{\ddot{T}}$, an $L_2$ loss of the velocity/acceleration of the 3D foot keypoints in global coordinates, and an $L_2$ loss on the global 3D body orientation $\boldsymbol{\tau}_i$. 
%See Sup.~Mat.~for details.

%% file: Sections/experiments.tex
\section{Experiments}
\label{sec:experiments}

\subsection{Implementation Details}

\textbf{Training details.}
During training, we directly use the ground truth trajectory of subjects to replace the estimated trajectory $\boldsymbol{\widehat{t}}$ for sampling the parameters.
We use the pre-trained backbone of BEV~\cite{bev} as the image backbone. 
We use RAFT~\cite{teed2020raft} as the optical flow backbone.
%to extract optical flow map. 
The training consists of two stages.
In the first stage, we freeze the weights of the backbones and train the head network for 40 epochs with a learning rate of 5e-5.
Then we train the image backbone and the head network together for 10 epochs with a learning rate of 1e-5.
We use four V100-16GB GPUs for training.
Limited by the GPU memory, we sample 4 video clips as a batch at each iteration; the clip length is 10 frames.

\textbf{Training and evaluation datasets.}
For training, we use three 3D human pose datasets (Human3.6M~\cite{h36m}, MPI-INF-3DHP~\cite{mono-3dhp2017,mehta2018single}, and 3DPW~\cite{3dpw}), two 2D human pose datasets (PennAction~\cite{zhang2013actemes} and PoseTrack~\cite{doering2022posetrack21}), and our DynaCam dataset. 
We evaluate TRACE on two multi-person in-the-wild benchmarks, 3DPW~\cite{3dpw} and MuPoTS-3D~\cite{mehta2018single}, and  DynaCam. 
%We evaluate global 3D trajectory estimation on DynaCam.
3DPW videos are most consistent with our tracking scenario.
%Videos in 3DPW dataset is captured by tracking subjects with a dynamic camera to record their daily activities, which highly match our task.
Unfortunately, not all 3DPW videos have complete tracking annotations. We select the 16 videos that do and call this subset Dyna3DPW.
We use this challenging subset to evaluate tracking and HPS accuracy in complex scenes with a moving camera.
%Therefore, we select 16 video sequences with complete tracking annotations from the 3DPW test set to form a Dyna3DPW subset for evaluating tracking in typical scenes. 
%The diverse activities in Dyna3DPW,
%e.g. shopping in toggery/market, running for bus, walking crowded bridge, working in office, 
%bring many challenges, such as person-person occlusion between the subjects or other non-tracked people, object occlusion, truncation, and sudden change of direction.
%The activities in Dyna3DPW include shopping in toggery/market, running for bus, going to a bar, having coffee, walking crowded bridge, working in office, etc. There are many challenging scenes in Dyna3DPW, such as pedestrians walking past the camera, occlusion between tracking subjects, object occlusion, truncation, and sudden change of direction.

\textbf{Evaluation metrics.}
For global 3D trajectory estimation, we compute the absolute trajectory error (\textbf{ATE})~\cite{grupp2017evo}
%average 3D position (\textbf{Traj}) and velocity (\textbf{Velocity}) error 
of the predicted global 3D trajectory $\boldsymbol{T}$ and the ground-truth after aligning with a similarity transformation. 
% like paying the taxes of doing tracking, can't imagine a paper for evaluation metrics.
For multi-object tracking, we report the ID switch (\textbf{IDs}), Multi-Object Tracking Accuracy (\textbf{MOTA}~\cite{bernardin2008evaluating}), Identification F1-score (\textbf{IDF1}~\cite{ristani2016performance}), and Higher Order Tracking Accuracy (\textbf{HOTA}~\cite{luiten2021hota}).
To assess the accuracy of 3D human pose/shape estimation, we compute the Mean Per Joint Position Error (\textbf{MPJPE}), Procrustes-aligned MPJPE (\textbf{PMPJPE}), and Mean Vertex Error (\textbf{MVE}).

Please refer to Sup.~Mat.~for more details.

\begin{table}[t]
\footnotesize
  \centering
    \resizebox{\columnwidth}{!}{
    \begin{tabular}{l|c|c|c}
	\toprule
        Method  & Translating Camera & Rotating Camera & FPS \\
        \midrule
        BEV~\cite{bev}+DPVO~\cite{teed2022deep} & 0.411 & 0.559 & 5.7 \\
        GLAMR~\cite{yuan2022glamr} & 0.912 & 0.816 & 1.5\\
        %BEV~~~~~+DPVO~~~~ & 0.411 & 0.559 & 5.7 \\
        %GLAMR & 0.912 & 0.816 & 1.5\\
        TRACE & \textbf{0.334} & \textbf{0.475} & \textbf{21.2}\\
    \bottomrule
    \end{tabular}}\vspace{-0.1in}
    \caption{Errors (ATE in $m$) in estimated 3D human trajectory in world coordinates and runtime efficiency on the DynaCam dataset.  }\label{tab:DynaCam}
\end{table}

\begin{table}[t]
\footnotesize
\centering
	\begin{tabular}{l|cccc} % p{1.7cm}
	\toprule
	Methods & IDs$\downarrow$ & MOTA$\uparrow$ & IDF1$\uparrow$ & HOTA$\uparrow$\\
    \midrule
    Trackformer~\cite{meinhardt2022trackformer} & 43 & 24.9 & 62.7 & 53.2\\ 
    Tracktor~\cite{bergmann2019tracking}  & 53 & 51.5 & 70.9 & 50.3 \\ 
    FlowPose~\cite{xiu2018poseflow} & 49 & 21.4 & 67.1 & 41.8\\ 
    T3DP~\cite{rajasegaran2021tracking} & 38 & 62.1 & 79.1 & 59.2 \\ 
    PHALP~\cite{rajasegaran2022tracking} & 22 & 66.2 & 81.4 & 59.4 \\ 
    ByteTrack~\cite{zhang2022bytetrack} & 15 & 73.3 & 84.6 & 63.5 \\
    TRACE & \textbf{0} & \textbf{86.9} & \textbf{93.4} & \textbf{65.3}\\ % 58 & 68.983 & 85.39 & 60.4
	\bottomrule
    \end{tabular}
    \vspace{-2mm}
    \caption{Tracking results on MuPoTS-3D~\cite{mehta2018single}. Results of the compared methods~\cite{meinhardt2022trackformer,bergmann2019tracking,xiu2018poseflow,rajasegaran2021tracking,rajasegaran2022tracking} are  from~\cite{rajasegaran2022tracking}.
    }\vspace{-2mm}
    \label{tab:mupots}
\end{table}

\begin{table}[t]
\footnotesize
\centering
	\begin{tabular}{l|cccc}
	\toprule
	Methods & IDs$\downarrow$ & MOTA$\uparrow$ & IDF1$\uparrow$ & HOTA$\uparrow$\\
    \midrule
    PHALP~\cite{rajasegaran2022tracking} & 5 & 97.9 & 93.7 & 74.2 \\ 
    YOLOX+ByteTrack & 3 & 97.3 & 98.2 & 73.1 \\
    BEV+ByteTrack & 37 & 93.6 & 79.1 & 59.3 \\
    %GLAMR~\cite{yuan2022glamr}\\
    TRACE w/o MO & 4 & 95.8 & 97.3 & 72.7\\
    TRACE & \textbf{1} & \textbf{99.3} & \textbf{99.7} & \textbf{74.7} \\
	\bottomrule
    \end{tabular}
    \vspace{-2mm}
    \caption{Tracking results on Dyna3DPW~\cite{3dpw}. w/o MO means ablating the estimated 3D motion offsets for tracking. 
    }\vspace{-2mm}
    \label{tab:dyna3dpw}
\end{table}

\begin{table}[t]
\footnotesize
	\centering
	\begin{tabular}{l|ccc}
	\toprule
{Method} & {PAMPJPE}$\downarrow$ & {MPJPE}$\downarrow$ & {PVE}$\downarrow$ \\
\midrule
HybrIK~\cite{li2021hybrik} & 45.0 & 74.1 & 86.5 \\
METRO~\cite{lin2021end} & 47.9 & 77.1 & 88.2 \\
%PHALP & 46.9 & 78.5 & 92.3 \\
GLAMR~\cite{yuan2022glamr} & 50.7 & - & - \\
CLIFF~\cite{li2022cliff} & \textbf{43.0} & \textbf{69.0} & \textbf{81.2}\\
D\&D~\cite{li2022dnd} & 42.7 & 73.7 & 88.6 \\
\midrule
ROMP~\cite{romp} & 47.3 & 76.7 & 93.4 \\
BEV~\cite{bev} & 46.9 & 78.5 & 92.3 \\
TRACE$^{\dag}$ & 50.8 & 80.3 & 98.1 \\ 
\bottomrule
\end{tabular}
\vspace{-0.1in}
	\caption{Comparisons with SOTA methods for 3D human pose and shape estimation on the 3DPW test set. ${\dag}$Please note that TRACE's results in original paper were wrong; see Sec.~\ref{sec:erratum}.}
	\label{tab:3DPW}\vspace{-2mm}
\end{table}

\begin{figure*}[t]
	\centerline{	\includegraphics[width=1\textwidth]{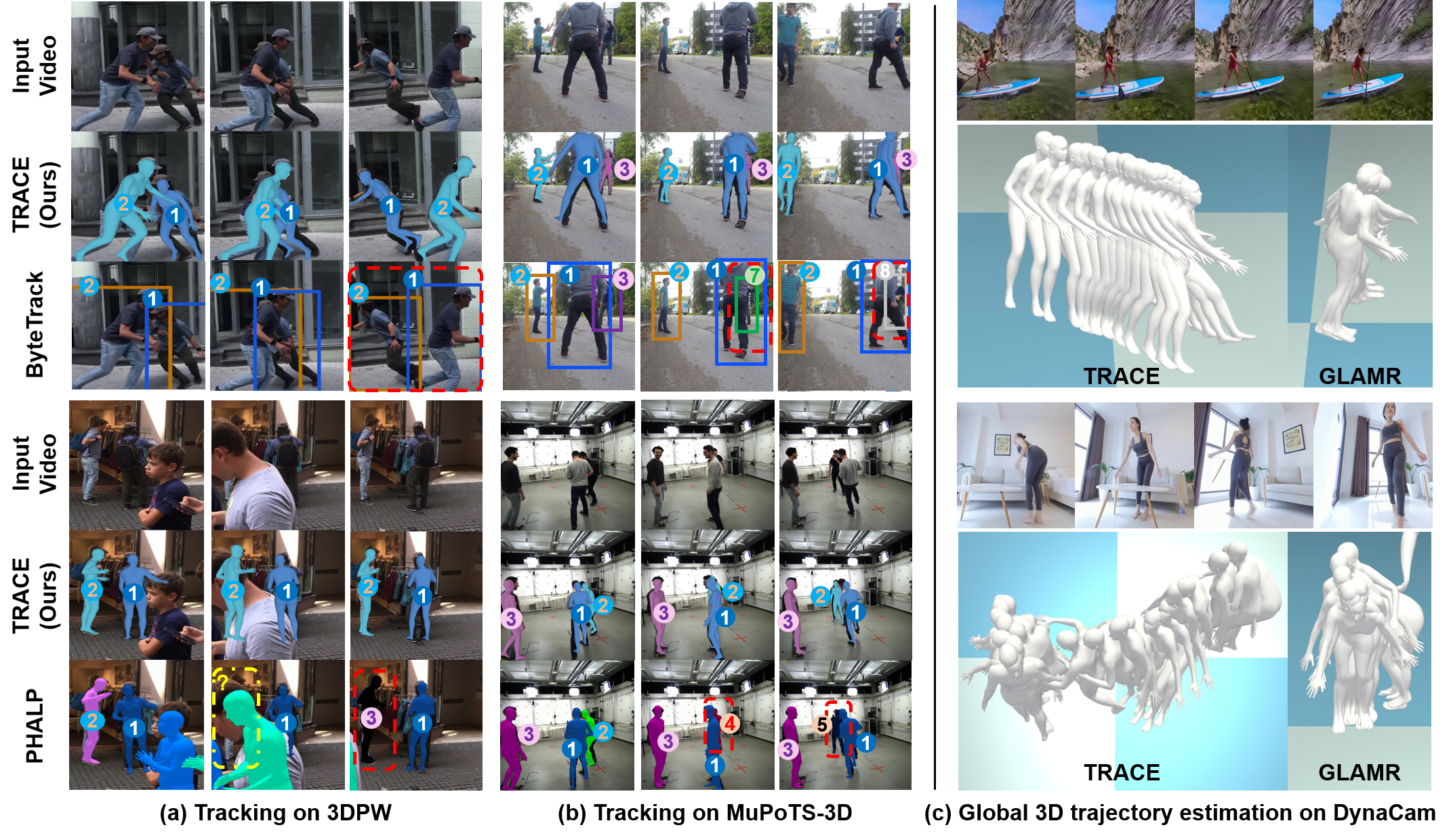}}
	\vspace{-0.1in}
	\caption{Qualitative comparisons to SOTA methods (ByteTrack~\cite{zhang2022bytetrack}, PHALP~\cite{rajasegaran2022tracking}, and GLAMR~\cite{yuan2022glamr}) on 3DPW, MuPoTS-3D, and DynaCam. Red dash boxes highlight the ID switches. Yellow dot-dash boxes highlight the misses under occlusions. }
	\label{fig:qualitative_results}\vspace{-1mm}
\end{figure*}

\subsection{Comparisons to State-of-the-art Methods}

\textbf{Global 3D trajectory estimation.} 
We aim to estimate the global human trajectory from dynamic cameras.
We do not explicitly estimate the camera motion. Instead, we use a world motion map to represent the global trajectory in world coordinates, which implies the camera motion.
Therefore, we evaluate the global trajectory error, instead of the camera pose, in Tab.~\ref{tab:dyna3dpw}.
First, we evaluate global 3D trajectory estimation on DynaCam.
%Since the most relevant method \cite{liu20214d} is not open-source, 
We compare TRACE with two baseline solutions.
The first one uses BEV to estimate the subjects in camera coordinates and DPVO \cite{teed2022deep}, a SLAM method, to estimate the camera and its motion; we call this BEV+DPVO.
%The most straightforward way is to directly estimate the camera poses from DC-videos.
%Since the previous method~\cite{liu20214d} is not open-sourced, we implement a baseline method, BEV+DPVO, by combining the BEV \cite{bev} for HPS and DPVO \cite{teed2022deep} for SLAM. 
As shown in Tab.~\ref{tab:DynaCam}, TRACE significantly outperforms BEV+DPVO in the accuracy of global 3D trajectory estimation.
The moving people in the scene make it hard for a SLAM method 
%Disturbed by the moving people in the scene, it is hard for SLAM method 
to extract stable corresponding keypoints.
Additionally, our synthetic camera motions differ from real camera motions and this may hurt DPVO's performance.
%This leads to the unsatisfactory results for BEV+DPVO, especially in cases where camera is rotating. 
A more direct comparison is to GLAMR~\cite{yuan2022glamr}.
%We also compare with the previous SOTA method, GLAMR~\cite{yuan2022glamr}, on DynaCam.
TRACE outperforms GLAMR in both accuracy and efficiency. 
In Fig.~\ref{fig:qualitative_results}(c), we also perform visual comparisons with GLAMR on DynaCam.
These results demonstrate the benefit of estimating global human trajectory using a holistic 5D representation.  
We provide more results in the supplemental video.

\textbf{Multi-subject tracking.}
To evaluate the performance of tracking subjects with dynamic cameras in real-world scenes, we compare TRACE with recent methods on Dyna3DPW.
PHALP~\cite{rajasegaran2022tracking} is a recent (SOTA) method that uses 3D cues and appearance to track people using SMPL. 
YOLOX+ByteTrack~\cite{ge2021yolox,zhang2022bytetrack} is a recently proposed and popular tracking-by-detection solution. 
These methods are designed to track all the people in a scene. 
Therefore, we process their results to avoid them being penalized for tracking unlabeled passers-by; 3DPW has annotations for at most 2 people in a scene but some scenes contain many people.
We first obtain their tracking results using their official code.
We then select the tracking results that achieve maximum IoU 
with the labeled ground truth subjects; we use these tracks for evaluation.
Note that, for a fair comparison with ByteTrack, TRACE runs in an on-line mode, without optimizing the past results. 
As shown in Tab.~\ref{tab:dyna3dpw}, TRACE outperforms PHALP, the previous 3D-representation-based method~\cite{rajasegaran2022tracking}, and tracking-by-detection~\cite{ge2021yolox,zhang2022bytetrack} methods on Dyna3DPW.
To evaluate the tracking robustness under long-term occlusion, we evaluate TRACE on MuPoTS-3D. 
Results of previous SOTA methods~\cite{meinhardt2022trackformer,bergmann2019tracking,xiu2018poseflow,rajasegaran2021tracking,rajasegaran2022tracking} are  from~\cite{rajasegaran2022tracking}.
Again, for a fair comparison, we filter out the unlabeled people in the ByteTrack results. 
As shown in Tab.~\ref{tab:mupots}, TRACE significantly outperforms previous SOTA methods. 
In particular, TRACE significantly reduces ID switches under long-term occlusions. 
These results illustrate the effectiveness and robustness of the proposed method for in-the-wild videos. 
The qualitative comparisons are presented in Fig.~\ref{fig:qualitative_results} and the supplemental video.

\textbf{3D human pose and shape estimation.}
Finally, we evaluate 3D human regression performance in DC-video using the 3DPW test set.
Because 3DPW does not provide ground-truth 3D translation in world coordinates, we evaluate root-relative 3D pose.
We compare TRACE with the recent one-stage~\cite{romp,bev} and multi-stage~\cite{li2021hybrik,lin2021end,yuan2022glamr,li2022cliff,li2022dnd} methods.

\vspace{-1mm}
\subsection{Ablation Studies}\vspace{-1mm}

\textbf{Temporal 5D representation v.s.~image-level 3D representation.}
%Our 3D detection and mesh regression are inspired by BEV~\cite{bev}. 
We go beyond BEV's image-level 3D representation and build a temporal 5D representation. 
As shown in Tab.~\ref{tab:DynaCam} and \ref{tab:dyna3dpw}, TRACE outperforms BEV or the multi-stage solutions using BEV on most metrics.
This demonstrates the value of learning a holistic 5D representation. 

\textbf{3D Motion Offset map.}
We also evaluate the effect of using predicted 3D motion offsets for tracking. 
As shown in Tab.~\ref{tab:dyna3dpw},  3D motion offsets improve performance by 3.5\%, 2.4\%, and 2.7\% in terms of MOTA, IDF1, and HOTA. 

%% file: Sections/conclusions.tex
\section{Conclusions}
\vspace{-1mm}

Human pose and shape estimation is not an end to itself.
Rather, estimating the 3D human in motion is useful for many tasks from behavior analysis to computer graphics.
However, to be useful, it is important to know the motion of humans with respect to the 3D scene and other people.
This means that HPS methods must estimate humans in a global  coordinate system and provide consistent tracks of people across time.
For generality, they also need to be able to do this from arbitrary moving cameras.
%These are hard problems that the field has only begun to address.

To tackle these challenging problems, we propose a novel 5D representation and a new neural architecture that reasons about people in 5D; that is, their 3D position, temporal trajectory, and identity.
Moving to a 5D representation enables our method, TRACE, to take a holistic view of the video, processing full frames and incorporating temporal features.
The core innovation of TRACE lies in its novel temporal representation in the form of new ``maps" that represent the motion of people across time in the camera and global coordinates.
These allow TRACE to be trained end-to-end, thus exploiting rich information from the video to solve the task.
TRACE is the first such single-shot method for 3D HPS estimation from video. %and it achieves SOTA results on common benchmarks.
%To help advance this line of research, the code and data will be publicly available.

\textcolor{\hcolor}{
Future work should look at explicitly estimating the camera, using training data like BEDLAM \cite{BEDLAM}, which contains complex human motion, 3D scenes, and camera motions.
We believe that camera motion and human motion provide complimentary information that can be used to recover human motion in world coordinates with metric accuracy.
}

%\textbf{Future work.}
%The setting of our task is just track the 3D motion and trajectory of assigned subjects. 
%Tracking and recovering the global 3D trajectory of all people in the scene would be a promising direction to make further exploration. 

%% file: Sections/acknowledgements.tex
\smallskip
\noindent \textbf{Acknowledgements:} This work was partially supported by the National Key R\&D Program of China under 
Grant (No. 2020AAA0103800) and Beijing Nova Program (No. 20220484063).

\smallskip
\noindent \textbf{MJB Disclosure:}
\url{https://files.is.tue.mpg.de/black/CoI_CVPR_2023.txt}

%\noindent \textbf{Disclosure:} MJB has received research gift funds from Adobe, Intel, Nvidia, Meta/Facebook, and Amazon.  MJB has financial interests in Amazon, Datagen Technologies, and Meshcapade GmbH.  While MJB is a consultant for Meshcapade, his research in this project was performed solely at, and funded solely by, the Max Planck Society.

%% file: Sections/errata.tex
\section{Erratum~\label{sec:erratum}}

\smallskip
%\noindent This section contains the errata in the original paper.

\smallskip
%\noindent \textbf{Errata 1:}
TRACE's results in Tab.~4 were wrong in the original version of the paper.
They were reported as:

\begin{table}[h!]
\footnotesize
\centering
\begin{tabular}{l|ccc}
\toprule
{Method} & {PAMPJPE}$\downarrow$ & {MPJPE}$\downarrow$ & {PVE}$\downarrow$ \\
\midrule
TRACE & 37.8 & 79.1 & 97.3 \\ 
\bottomrule
\end{tabular}
\end{table} 

\smallskip
\noindent and have been corrected to 

\begin{table}[h!]
\footnotesize
\centering
\begin{tabular}{l|ccc}
\toprule
{Method} & {PAMPJPE}$\downarrow$ & {MPJPE}$\downarrow$ & {PVE}$\downarrow$ \\
\midrule
TRACE & 50.8 & 80.3 & 98.1 \\ 
\bottomrule
\end{tabular}
\end{table} 

\smallskip
\noindent 
The computation of the per-frame PA-MPJPE was performed incorrectly in a widely used PA-MPJPE evaluation function. A bug resulted in the incorrect shape of the 3D joints for Procrustes Analysis (PA) when the annotations were for two people. 
In such cases, PA would generate a joint-wise $14x14$ transformation matrix for aligning the 3D joints of shape $14x3$, instead of the original $3x3$ rotation matrix we expected. 
%During computing frame-by-frame PA-MPJPE, a code bug results in the wrong input format of 3D joints when there are annotations for two people. We sincerely apologize for this error. 

This error occurs in the widely used PA-MPJPE evaluation function when the 0-dimensional shape of the joint matrix is equal to 2 or 3. 
If you use the same code, please consider fixing this issue in your evaluation code.
For details, please refer to \url{https://github.com/Arthur151/ROMP/blob/master/simple_romp/trace2/PMPJPE_BUG_REPORT.md}. 